\documentclass[12pt]{article}

\usepackage{arxiv}

\usepackage[utf8]{inputenc}
\usepackage[T1]{fontenc}
\usepackage{hyperref}
\usepackage{url}
\usepackage{booktabs}
\usepackage{amsfonts}
\usepackage{graphicx}
\usepackage{subfigure}
\usepackage{multirow}
\usepackage[numbers]{natbib}
\usepackage{enumitem}
\usepackage{array}

\newcolumntype{C}[1]{>{\centering\let\newline\\\arraybackslash\hspace{0pt}}m{#1}}

\newcommand{\eqref}[1]{(\ref*{#1})}

\title{Explainable Image Similarity: Integrating Siamese Networks and Grad-CAM}

\date{}                    % Or removing it

\author{ \href{https://orcid.org/0000-0002-3996-3301}{\includegraphics[scale=0.06]{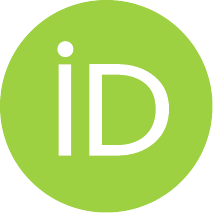}\hspace{1mm}Ioannis E.~Livieris}\thanks{Corresponding author}\\
    Department of Statistics and Insurance,\\
    University of Piraeus,\\
    Piraeus, GR 185-34.\\
    \texttt{livieris@unipi.gr} \\
    \And
	\href{https://orcid.org/0000-0002-3562-461X}{\includegraphics[scale=0.06]{orcid.pdf}\hspace{1mm}Emmanuel Pintelas}\\
	Department of Mathematics,\\
	University of Patras,\\
	Patras, GR 265-00.\\
	\texttt{livieris@upatras.gr} 
    \And
	\href{https://orcid.org/0000-0003-1729-4124}{\includegraphics[scale=0.06]{orcid.pdf}\hspace{1mm}Niki Kiriakidou}\\
	Department of Informatics and Telematics,\\
	Harokopio University of Athens,\\
	Athens, GR 177-78.\\
	\texttt{kiriakidou@hua.gr} \\  	
    \And
	\href{https://orcid.org/0000-0001-8436-2743X}{\includegraphics[scale=0.06]{orcid.pdf}\hspace{1mm}Panagiotis Pintelas}\\
	Department of Mathematics,\\
	University of Patras,\\
	Patras, GR 26500.\\
	\texttt{ppintelas@gmail.com} \\  		
}

\renewcommand{\shorttitle}{Explainable Image Similarity: Integrating Siamese Networks and Grad-CAM}

\hypersetup{
    pdftitle={Explainable Image Similarity: Integrating Siamese Networks and Grad-CAM},
    pdfsubject={cs.CV, cs.LG, cs.AI},
    pdfauthor={Ioannis E. Livieris, Emmanuel Pintelas, Niki Kiriakidou and Panagiotis Pintelas},
    pdfkeywords={Explainability, Siamese networks, Grad-CAM, recommendations},
}

\begin{document}
    \maketitle

\begin{abstract}
	With the proliferation of image-based applications in various domains, the need for accurate and interpretable image similarity measures has become increasingly critical. Existing image similarity models often lack transparency, making it challenging to understand the reasons why two images are considered similar. In this paper, we propose the concept of explainable image similarity, where the goal is the development of an approach, which is capable of providing similarity scores along with visual factual and counterfactual explanations. Along this line, we present a new framework, which integrates Siamese Networks and Grad-CAM for providing explainable image similarity and discuss the potential benefits and challenges of adopting this approach. In addition, we provide a comprehensive discussion about factual and counterfactual explanations provided by the proposed framework for assisting decision making. The proposed approach has the potential to enhance the interpretability, trustworthiness and user acceptance of image-based systems in real-world image similarity applications. The implementation code can be found in \url{https://github.com/ioannislivieris/Grad_CAM_Siamese.git}.\\ \\
	*** This paper has been accepted for publication at \textit{Journal of Imaging}. Cite: Livieris, I. E., Pintelas E., Kiriakidou, N., \& Pintelas, P. (2023). Explainable Image Similarity: Integrating Siamese Networks and Grad-CAM. \textit{Journal of Imaging}, 9(10):224.***
\end{abstract}

% keywords can be removed

\section{Introduction}

In many real-world scenarios, the ability to measure image similarity is crucial for decision-making processes, intelligent systems as well as user interactions; therefore, image similarity models constitute a vital role in various computer vision tasks \cite{bell2016inside,gordo2017end,gygli2014creating,shin2016deep, radenovic2018fine}. For example, in image retrieval systems, users often search for similar images based on a reference image or specific visual features \cite{gordo2017end}. Image similarity models allow these systems to find relevant images quickly and accurately. In content-based image analysis, large image databases are categorized and organized using image similarity models; hence, enabling the efficient automatic identification of similar images \cite{bell2016inside}. In copyright infringement detection or multimedia management, image similarity assists in identifying duplicate or visually similar images \cite{gygli2014creating}. Furthermore, in medical imaging, comparing and matching medical images can aid in the diagnosis and identification of diseases or abnormalities \cite{shin2016deep}. Finally, image similarity can also assist in visual search engines, where users are able to visually find similar images without relying on text-based queries \cite{radenovic2018fine}.

Siamese neural networks \cite{chicco2021Siamese}, probably constitute the most efficient and widely utilized class of image similarity models. During the last decade, they have been successfully applied for addressing image similarity tasks by quantifying the similarity between images through numerical values \cite{appalaraju2017image,melekhov2016Siamese,rossi2020multi}. The backbone of this class of neural networks are convolutional layers which  are characterized by their remarkable abilities for image processing.
Nevertheless, due to their architectural design,  Siamese networks are not able to provide the users with human-understandable explanations about why two images are deemed similar. As the adoption of image-based technologies continues to grow in diverse applications like medical imaging, e-commerce, social media and security, the need for explainability in image similarity becomes paramount \cite{selbst2018intuitive}.
Explainability is a critical aspect of Deep Learning (DL), especially when dealing with complex models composed by convolutional layers. Although that convolutional-based neural network models, such as  Siamese networks, are highly effective in several image processing tasks, they lack in transparency and explainability; thus, they are considered as ``black boxes'' \cite{pintelas2020explainable}. 
Notice that many traditional machine learning models, such as decision trees and linear models, often have the advantage of being interpretable, since their decision-making process is based on understandable features; nevertheless, Siamese networks learn intricate and abstract features through layers of convolutions, making it challenging to directly interpret their decisions.

Explainability techniques aim to shed light by providing insights into how and why a convolutional-based model makes certain predictions by understanding the features and patterns, which the model learns from the data. These techniques not only enhance our understanding about the model's decision process but also play a vital role in building trust and accountability in artificial intelligence systems. More specifically, they enable us to verify the reasoning behind the predictions \cite{ribeiro2016should}, identify potential biases, errors, or misinterpretations in model predictions and provide a means to improve their performance \cite{lundberg2017unified}. Also, in some domains, there are strict regulations that require models to be interpretable. For instance, the General Data Protection Regulation in Europe includes the "\textit{right to explanation}", which mandates that individuals should be provided with an explanation for automated decisions \cite{wachter2017transparent}. Finally, in certain contexts, there may be legal or ethical requirements to explain model predictions to end-users or stake-holders, making interpretability a crucial aspect of the deployment \cite{livieris2023advanced,selbst2018intuitive}.

In the literature, several research directions have focused on enhancing the interpretability of deep learning models, particularly in the fields of computer vision \cite{pintelas2021novel,samek2017explainable}. Explainable artificial intelligence (XAI) techniques, such as attention mechanisms \cite{woo2018cbam} and Gradient-weighted Class Activation Mapping (Grad-CAM) technique \cite{selvaraju2017grad}, have been successfully applied to image classification, object detection and semantic segmentation tasks. However, the application of XAI to image similarity remains underexplored. In light of the increasing adoption of image-based technologies across various domains, the demand for explainable image similarity is considered crucial. Users and decision-makers seek transparency in understanding why certain images are considered similar, especially in critical applications like medical diagnosis or security surveillance. Therefore, exploring the integration of new or existing XAI techniques \cite{arrieta2020explainable} with image similarity models \cite{ma2021image} is able to provide insights on the underlying similarities between images. Moreover, exploring the notion of similarity from a human-centric perspective may lead to novel contributions in image understanding and user-friendly applications.

In this work, we propose a new concept, named ``\textit{explainable image similarity}''. Our primary aim is to bridge the gap between numerical similarity scores and human-understandable explanations. Along this line, we propose a new algorithmic framework, which integrates Siamese Networks and Grad-CAM for providing explainability in image similarity tasks. The former are utilized for calculating the similarity between two input images while the latter is used for visualizing and interpreting the decisions made by convolutional-based Siamese network. An attractive advantage of the proposed framework is that it is able to provide image similarity score along with visual intuitive explanations for its decisions (factual explanations) together with explanations based on its ability to ``what if'' scenarios (counterfactual explanations). Finally, we provide a comprehensive discussion about factual and counterfactual explanations as well as the valuable insights and recommendations which can be made from the application of the proposed framework on three real-world use case scenarios.

At this point it is worth mentioning that although, Grad-CAM technique has been widely used and studied in a variety of domains to the best of our knowledge, it has never been utilized for image similarity tasks. 

\clearpage

\noindent{}Summarizing, the main contributions of this work are described as follows:
\begin{itemize}
	\item We propose the concept ``\textit{explainable image similarity}'' highlighting the needs for providing human-understandable explanations for image similarity tasks.
	
	\item We propose a new conceptual framework for explainable image similarity, which integrates Siamese networks along with Grad-CAM technique, which is able to provide reliable, transparent and interpretable decisions on image similarity tasks. 
	
	\item The proposed framework produces factual and counterfactual explanations, which are able to provide valuable insights and be used for making useful recommendations.
\end{itemize}

The rest of this paper is organized as follows: Section~\ref{Sec:2} presents the state-of-the-art works relative to Grad-Cam technique and image similarity applications. Section~\ref{Sec:3} presents the concept of ``\textit{explainable image similarity}'' as well as a detailed discussion about the proposed framework while Section~\ref{Sec:4} presents three use cases scenarios from its application. Finally, Section~\ref{Sec:5} discusses the proposed research, summarizes its conclusions and provides some interesting ideas for future work.

\section{Related work}\label{Sec:2}

Convolutional-based Neural Networks (CNNs) revolutionized modern computer vision and are widely regards as the cornerstone choice for addressing image processing tasks
\cite{pintelas2023multi,pintelas2021convolutional,radenovic2018fine,shin2016deep}. % LIVIERIS: cnn multi-view framework
%\cite{hassaballah2020deep,khan2018guide,shin2016deep,radenovic2018fine}. % LIVIERIS: cnn multi-view framework/
The core element of CNNs are convolutional layers, which exploit a set of learnable filters (kernels) for generating feature maps. The aim is to highlight distinct attributes like edges, textures and shapes, allowing subsequent layers to recognize higher-level representations. 

Nowadays, explainability and interpretability play a significant role in bridging the gap between the advanced capabilities of DL models and the need for transparency and accountability in their decision-making processes. However, as CNNs become deeper and more complex, understanding how and why they make particular predictions becomes challenging. Grad-CAM \cite{selvaraju2017grad} is a novel technique, which enhances the interpretability of CNNs focusing on highlighting the regions of an input image that significantly contribute to a specific prediction; thus, it has been applied in various applications.
Hsiao et al. \cite{hsiao2022application} exploited the flexibility of the Grad-CAM technique towards accurate visualization and interpretable explanation of CNNs. In particular, the authors utilized Grad-CAM to provide reliable and accurate analysis results for fingerprint recognition. Generally, fingerprints are difficult to be analyzed manually; hence, this study contributed to the assistance of criminal investigation cases. In a similar research, Sang-Ho et al. \cite{kim2023combining} provided another application of the Grad-CAM technique in which they focused on providing a trading strategy for simultaneously achieving higher returns, compared to benchmark strategies. 
Along this line, the authors used Grad-CAM technique in conjunction with a CNN model aiming to develop a trustworthy method for meeting explainability as well as profitability in finance, therefore, fulfilling the challenging investors' needs.

In computer vision, the concept of image similarity consists of a fundamental building block for various real-world applications, ranging from image retrieval \cite{singh2020explainable} and pattern recognition \cite{kim2023combining} to anomaly detection \cite{saeki2019visual}. Siamese networks \cite{chicco2021Siamese} have been established as state-of-the-art models for tackling image similarity tasks, especially where the available labeled data are limited. Their special architectural design enables them to learn and capture intricate relationships between pairs of images, allowing for the precise quantification of similarity and/or dissimilarity. 

Appalaraju and Chaoji \cite{appalaraju2017image} proposed a new approach for identifying similar images using a deep Siamese network, named SimNet. In more detail, SimNet is trained on pairs of positive and negative images using a novel online pair mining strategy (OPMS). OPMS has been inspired by curriculum learning, a methodology for training DL models, aiming to ensure consistently increasing difficulty of input image pairs during the training process. 
Furthermore, another characteristic of SimNet is that it is composed of a
multi-scale CNN, which is able to learn a joint image embedding of top and lower layers. 
For evaluating the model's performance, they utilized the widely used computer-vision object recognition dataset, named CIFAR10. 
The experimental analysis and use case examples showed that the proposed SimNet model
is able to better capture fine-grained similarities between images, compared to traditional CNNs. Additionally, the authors stated that the adopted curriculum learning strategy led to faster model training.

Melekhov et al. \cite{melekhov2016Siamese} proposed a novel methodology for exploiting Siamese networks for dealing with image similarity and classification problems.  For detecting the matching and non-matching image pairs, the authors suggested to represent them as feature vectors and distinguish the similarity between the input images using the Euclidean distance of these calculated feature vectors. 
In particular, those feature vectors are obtained through convolutional layers while 
the model training was based on contrastive loss.
In their research, the authors used a large set of images from five different landmark for evaluating the performance of the proposed Siamese model for image matching against widely used models such as AlexNet, HybridNet and sHybridNet. Based on their experimental analysis the authors concluded that the proposed model reported promising performance on image similarity and classification tasks while in contrast to traditional models, it is able to efficiently handle datasets with imperfect ground truth labels.

Rossi et al. \cite{rossi2020multi} introduced a novel supervised Siamese deep learning architecture, which is a new Content–Based Image Retrieval system (CBIR) for assisting the process of interpreting a prostate radiological 
Magnetic Resonance Image (MRI). The rationale behind the architecture of the proposed approach is to integrate all available information in multi–parametric medical imaging tasks for predicting diagnostically similar images. 
Additionally, for handling multi–modal and multi–view MRIs, the authors considered the diagnostic severity of the lesion, assessed by the PI-RADS score \cite{gupta2020pi}, as the similarity criterion. It is worth mentioning that despite its initial purpose of development, this approach can be utilized for several diagnostic medical imaging retrieval due to its general design. As regards the experimental analysis, the authors presented that the performance of Siamese-based CBIRs was superior to that of the most widely used autoencoder-based CBIRs, for both diagnostic and information retrieval metrics.

In this research, we introduce the concept of explainable image similarity, for providing useful, interpretable and transparent insights into the underlying factors driving image relationships and comparisons. In addition, we propose a new framework which integrates Siamese networks together with Grad-CAM technique. The former are used for calculating the similarity between input images while the latter is used for visualizing and interpreting the decisions made by convolutional-based neural networks. In contrast to previous presented state-of-the-art approaches, the proposed framework is able to provide image similarity score along with visual intuitive explanations for its decisions. The presented use case scenarios demonstrate the applicability of the proposed framework as well as a path for providing insights and useful recommendations from factual and counterfactual explanations.

\section{Explainable image similarity}\label{Sec:3}

In this section, we present the proposed framework which is able to provide similarity scores along with visual transparent and understandable explanations for its decisions. We recall that our primary goal is to proposed the concept of explainable image similarity for bridging the gap between numerical 
similarity scores and human-understandable explanations.
By offering interpretable explanations, explainable image similarity not only enhances the usability of similarity-based applications but also empowers users to comprehend the reasoning behind the model's decisions, ultimately fostering informed and confident decision-making.

In the following, we briefly present the main components of the proposed framework which is based on the integration of Grad-CAM technique to Siamese Networks
as well as a detailed description, paying special attention to its capabilities and advantages.

\subsection{Background}

\textit{Siamese neural networks} \cite{chicco2021Siamese} constitute a special class of deep learning architectures, which are used in tasks involving similarity comparison, such as image or text matching
\cite{appalaraju2017image,rossi2020multi,neculoiu2016learning}.
These networks are characterized by their robustness to data which exhibit variations, distortions or noise as well as their requirement of significantly less labeled training data compared to neural networks; therefore, they have been well-established for real-world scenarios \cite{singh2020explainable,kim2023combining,saeki2019visual,guo2023Siamese,mazzeo2020Siamese}. A traditional Siamese network is composed by two identical sub-networks with shared weights (backbone-network), allowing them to extract and encode into fixed-size feature vectors (embeddings) from input pairs. Then, the similarity
of the input images (similarity score) is provided by computing the distance between the calculated embeddings.

\textit{Gradient-weighted Class Activation Mapping} (Grad-CAM) \cite{selvaraju2017grad} is a powerful and model-agnostic technique in the field of computer vision, which enhances the interpretability of deep neural networks. Grad-CAM provides a way to visualize and localize the regions of an input image, which contribute most to the model's decision. For obtaining the class-discriminative localization map, denoted by $L_{Grad-CAM}$, we initially calculate the neuron importance weights $\alpha_k$ using the gradient of the model's output $y$ with respect to the $k$-th map activations $A^k$ of a selected convolutional layer, which are flowed back and are global-average-pooled over the width (index $i$) and height (index $j$) dimensions, that is
\begin{equation}\label{Eq:Neuron importance weights}
	\alpha_k = \frac{1}{Z} \sum_i \sum_j \frac{\partial y}{\partial A^k_{ij}},
\end{equation}
where $Z$ is the total number of spatial locations in the feature maps. Then, we perform a weighted combination of forward activation maps, and follow it by ReLU activation function for calculating $L_{Grad-CAM}$, namely
\begin{equation}\label{Eq:Grad-Cam}
	L_{Grad-CAM} = \textnormal{ReLU} \left( \sum_k a_k A^k \right).
\end{equation}
By utilizing the gradients with respect to the model's internal feature maps, Grad-CAM generates an activation map, which highlights the discriminative regions responsible for the model's decision.

\subsection{Proposed framework}

\begin{figure}[!ht]
	\centering
	\includegraphics[width=15cm]{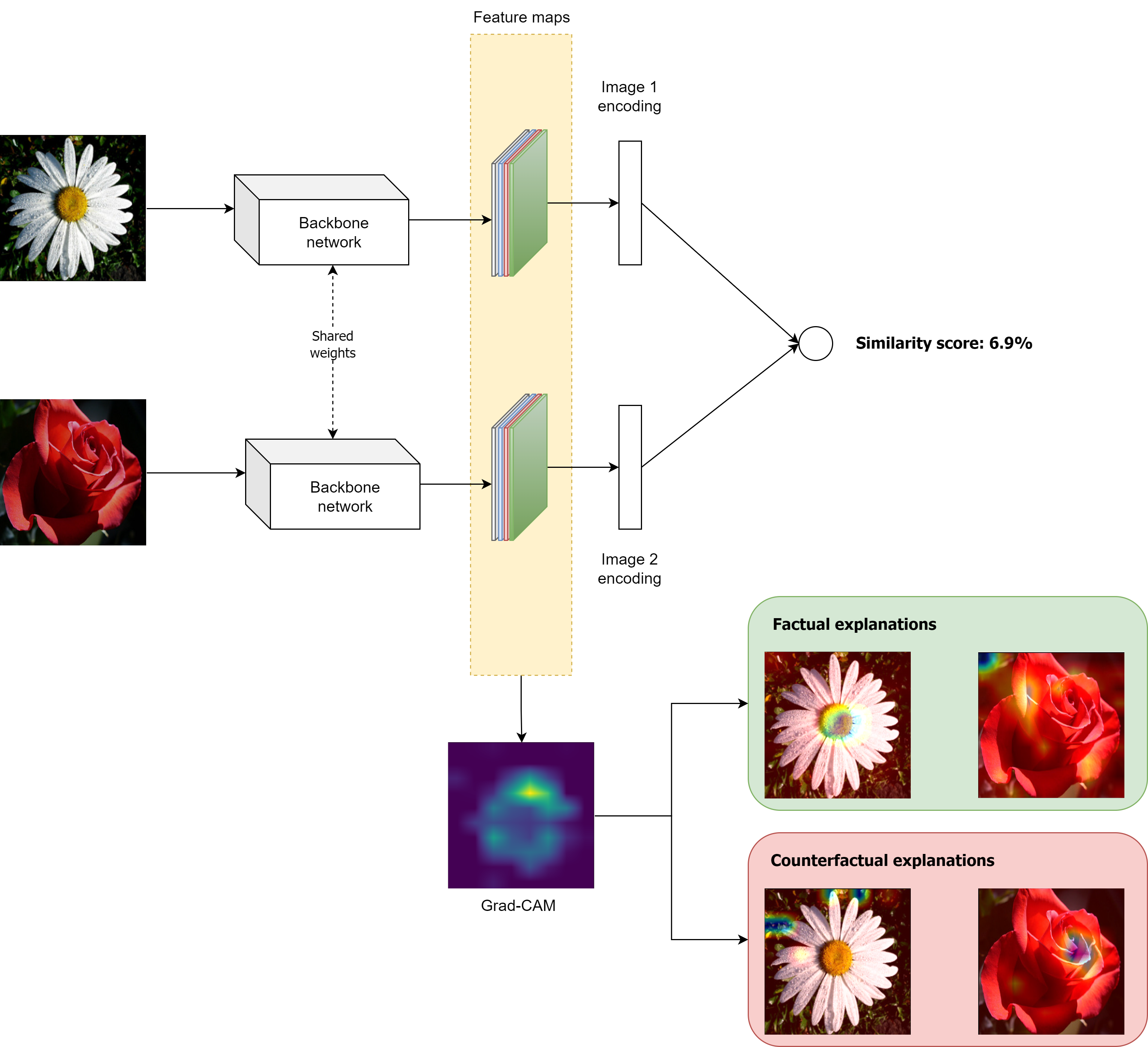}
	\caption{Architecture of the proposed framework\label{Fig:Architecture}}
\end{figure}

Next, we provide a detailed description of the proposed framework while a high-level presentation of its architecture is highlighted in Figure \ref{Fig:Architecture}.
Initially, two images are considered as input in a Siamese network, which are processed by the backbone network for encoding them into fixed-size feature vectors (embeddings). Then, the image embeddings are used for discerning similarities and differences between the input images and ultimately calculating their similarity score. Independently, Grad-CAM technique is applied to the last convolutional layer of the backbone network for the development of the Grad-CAM heatmaps and visualize the features, which significantly impact the Siamese model's decisions (factual explanations). 

In addition, the proposed framework is able to provide with counterfactual explanations. Actually, a counterfactual explanation provides a description of ``\textit{what would have not happened when a certain decision was taken}'' \cite{selvaraju2017grad}. This transparency not only enhances model's interpretability but also empowers stakeholders to identify potential biases, assess model fairness, and build trust in AI-driven systems, leading to more accountable and reliable artificial intelligence solutions. The counterfactual explanations can be easily developed by a slight modification to Grad-CAM technique, namely, by simply replacing $y$ with $1-y$ in Eq. \eqref{Eq:Neuron importance weights}.

\clearpage

\noindent{}Summarizing, the advantages of the proposed framework are:
\begin{itemize}
	\item \textit{Counterfactual explanations:} The identification of regions, which would make the network change its prediction, could highlight concepts that confuse the model. Therefore, by removing those concepts, the model's decisions may be more accurate or more confident.
	
	\item \textit{Bias evaluation of model's decisions:} In case, the Siamese model is performing well on both training and testing data (not-biased model), Grad-CAM heatmaps may be used to visualize the features, which significantly impact the model's decisions. In contrast, in case the Siamese model is performing pretty on the training data but it is not able to generalize well (biased model), Grad-CAM heatmaps can be efficiently used to identify unwanted features in which the model focuses on.
\end{itemize}

\section{Application of proposed framework and use case scenarios}\label{Sec:4}

Next, we provide some use case scenarios from the application of the proposed framework to three (3) well-known datasets from different real-world application domains:
\begin{itemize}
	\item \textit{Flowers}. This dataset contains 4242 images (320x240) of flowers, which were categorized in five classes: ``chamomile'', ``tulip'', ``rose'', ``sunflower'' and ``dandelion''. 
	\item \textit{Skin cancer}. This dataset concerns images (224x224) of 1400 malignant and 1400 benign oncological diseases.

	\item \textit{AirBnB}. This few-show dataset is composed by 864 interior and exterior house pictures (600x400) scraped from AirBnB over three cities, % (Boston, Berlin, Seattle) 
	which were classified in 12 classes: ``backyard'', ``basement'', ``bathroom'', ``bedroom'', ``decor'', ``dining-room'', ``entrance'', ``house-exterior'', ``kitchen'', ``living-room'', ``outdoor'', ``staircase'' and ``tv-room''.
\end{itemize}

The presented use cases focus on highlighting how the proposed framework could be used for image similarity tasks, what useful conclusions could be drawn by factual and counterfactual explanations and finally, what useful recommendations could be provided.

\noindent{}For training the Siamese networks, 80\% of each dataset's images were used for training while the rest 20\% for testing while preserving the variance of each class in each set. In addition, 10\% of training images where used as validation set for optimizing the network's performance. The implementation code along with the datasets can be found in \url{https://github.com/ioannislivieris/Grad_CAM_Siamese.git}.

Based on the images of each training dataset, we created the training pairs as follows: for each image, two images were randomly selected; one image from the same class and another image from a different class. The first pair containing the images from the same class is assigned with label zero (0), while the 
second pair containing the images from different classes is assigned with label one (1). Along this line, the similarity between two random input images is defined by $1-d$, where $d$ is the Siamese model's output. Notice that this methodology was initially proposed by Melekhov et al. \cite{melekhov2016Siamese}.

At this point, it is worth mentioning that the model's prediction can be exploited to obtain information if two images belong to the same class or not. More specifically if the prediction of the Siamese network for a pair of images is less than a pre-defined $threshold$, then the images are considered similar (belong to the same class); otherwise, they are considered as dissimilar (belong to the different classes). Notice that in our experiments, $threshold$ was set to 0.5.

As regards, the Siamese network architecture: ResNet50 \cite{he2016deep} was used as backbone network, followed by average pooling layer of size $(1,1)$ and a dense layer of 256 neurons with ReLU activations for calculating each input image embeddings. Next, the $L_2$-distance between the embeddings is calculated, followed by an output layer with one neuron with Sigmoid activation function. 
The utilized architecture and hyperparameter selection provide us very good and reliable performance as regards all three benchmarks. It is worth highlighting the scope of this research was not to address a specific class of benchmarks i.e. few-shot learning benchmarks, one-shot learning benchmarks, etc neither to provide a new advanced model architecture but to provide human-meaningful explanations on similarity tasks though the proposed framework.
Finally, the Siamese model was trained using ADAM algorithm \cite{reddi2019convergence} while contrastive loss function \cite{wang2021understanding} was used for training the network, which is defined by
$$
\mathcal{L}=\frac{1}{2} \left[ (1-y)(D_w)^2 + y	\{max(0, m-D_w)\}^2 \right],
$$
where $D_w$ is the model's output and $m$ is the margin value, which was set to 2.

\subsection{Flowers dataset}

\begin{figure}[!ht]
	\centering
	\subfigure[]{\includegraphics[width=4.5cm]{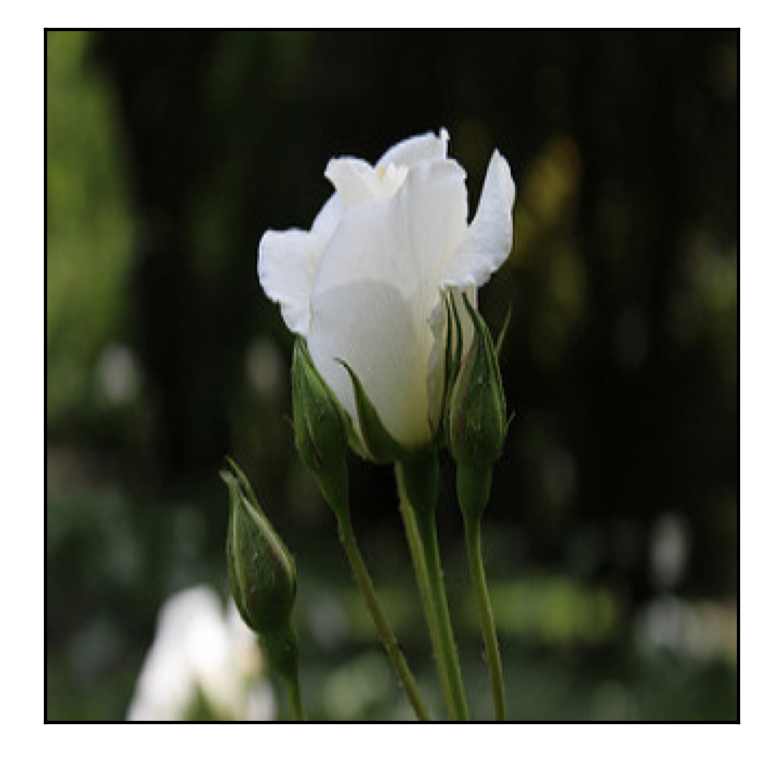}}
	\subfigure[]{\includegraphics[width=4.5cm]{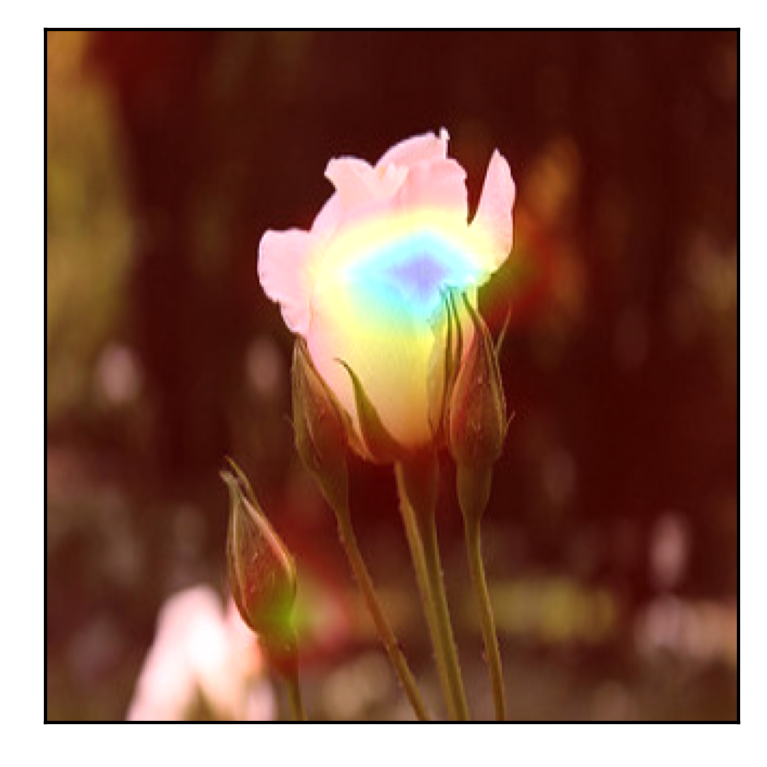}}
	\subfigure[]{\includegraphics[width=4.5cm]{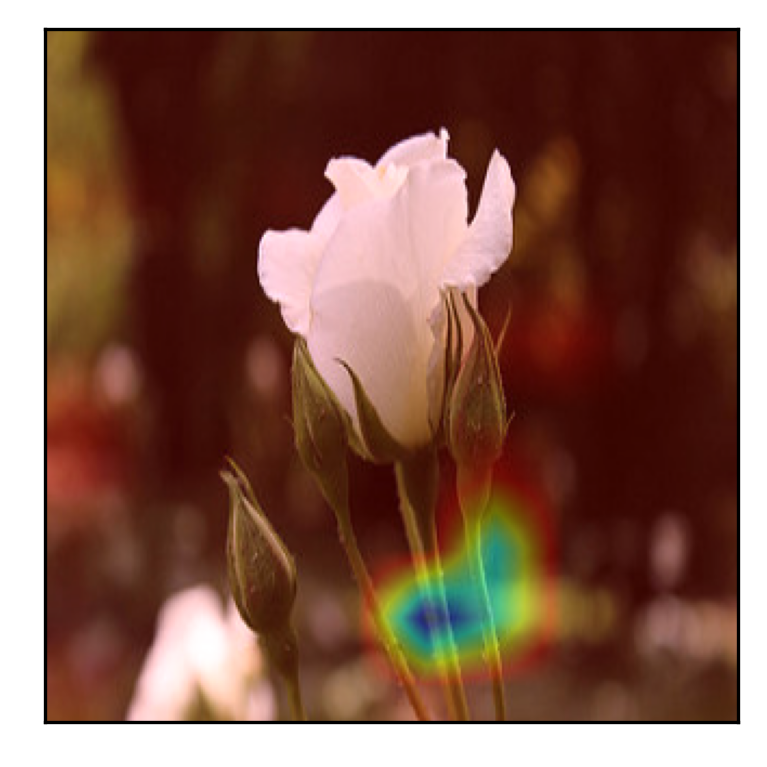}}\\
	\subfigure[]{\includegraphics[width=4.5cm]{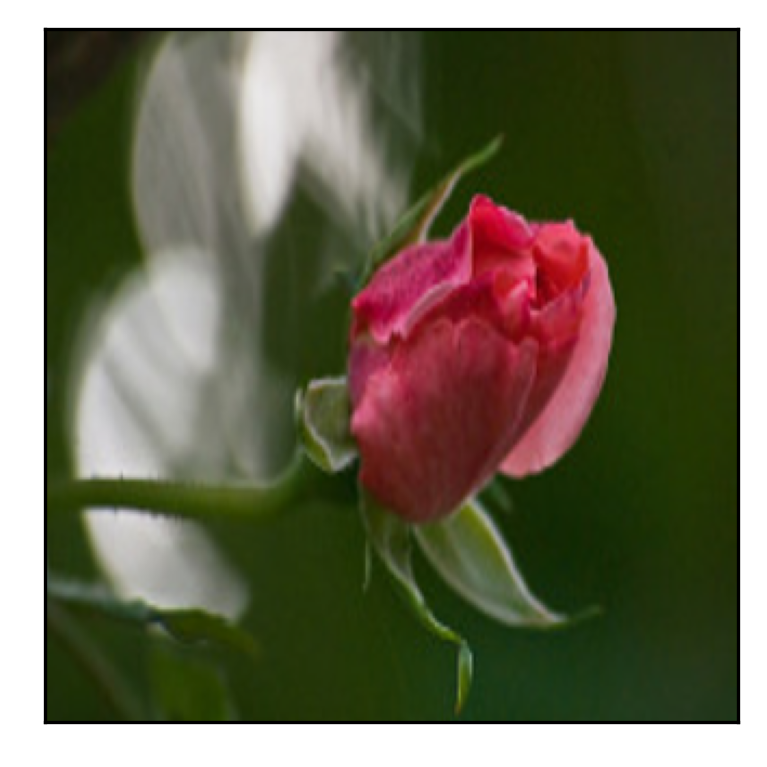}}
	\subfigure[]{\includegraphics[width=4.5cm]{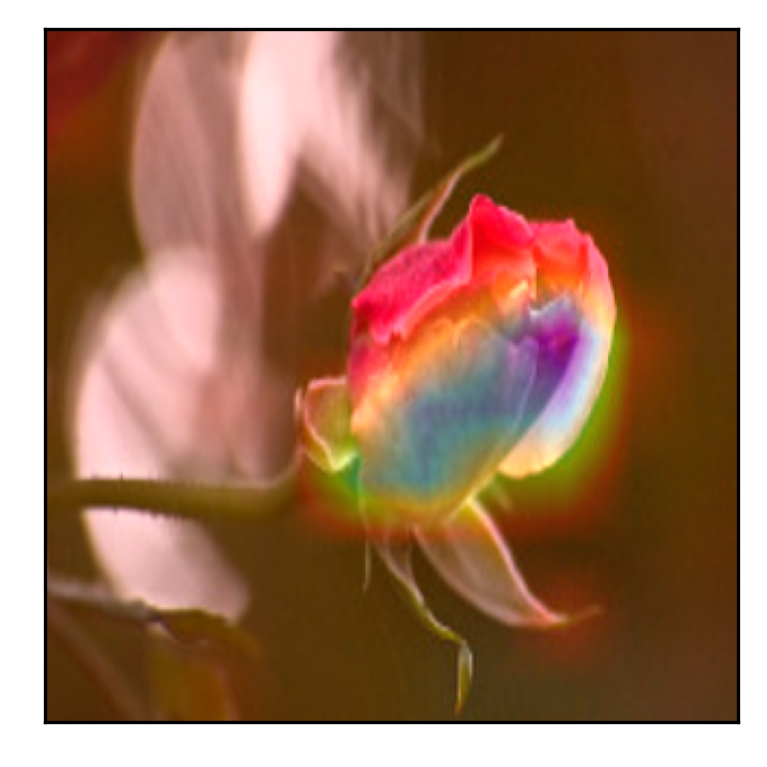}}
	\subfigure[]{\includegraphics[width=4.5cm]{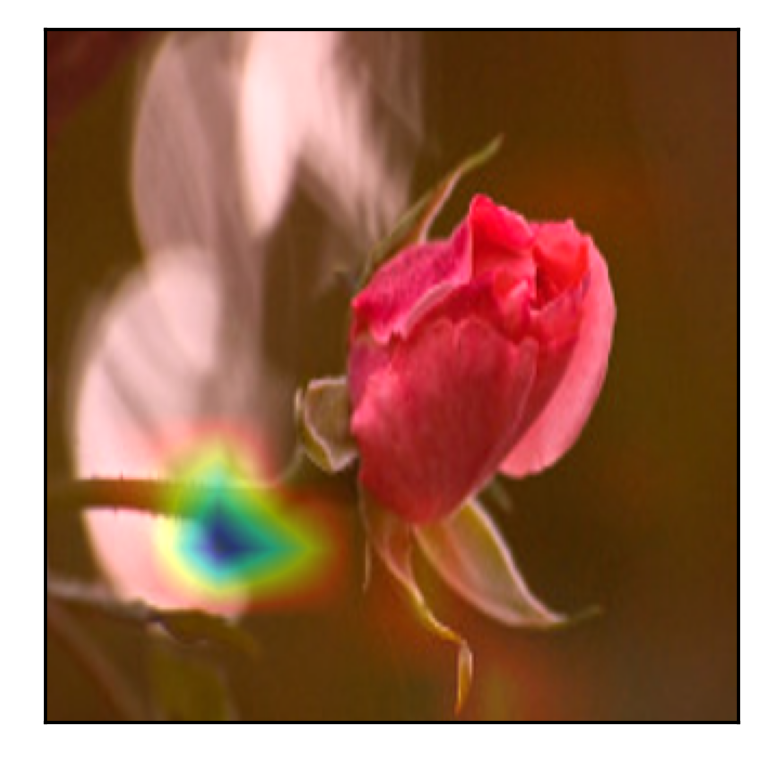}}		
	\caption{Application of the proposed framework on flowers dataset (a) Original input image$_1$ (b) Factual explanations on image$_1$ (c) Counterfactual explanations on image$_1$ (d) Original input image$_2$ (e) Factual explanations on image$_2$ (f) Counterfactual explanations on image$_2$}\label{Fig:Flower use case}
\end{figure}

Next, we present an example from the application of the proposed framework on two random images (Figure \ref{Fig:Flower use case}(a) and \ref{Fig:Flower use case}(d)) from Flowers dataset, which belong to the same class (``rose''). The Siamese model's prediction was 0.24, which implies that the model predicts that 
the similarity between the input images is 76\%. In addition, since the similarity score is greater than the pre-defined $threshold=0.5$, the model suggests that input images belong to the same class. 
Figures \ref{Fig:Flower use case}(b) and \ref{Fig:Flower use case}(e) present the factual explanations provided by Grad-CAM in order to identify the features, which impact the model's decisions. In more detail, the model's decision was based on flower's blossoms in which it found common characteristics. As regards, the counterfactual explanations, which are presented in Figures \ref{Fig:Flower use case}(c) and \ref{Fig:Flower use case}(f), they highlight that the model would have been based on the stems of  both flowers for predicting that the images are not similar. 

By taking into consideration that similar conclusions can be drawn by randomly selecting any pair of images in Flowers dataset, a possible recommendation for improving the model's performance could be that the model is based on identifying the blossoms in the input images for making its prediction; thus, a removal of other characteristics such as stems, background, etc may improve the model's performance.

\subsection{Skin cancer dataset}

\begin{figure}[!ht]
	\centering
	\subfigure[]{\includegraphics[width=4.5cm]{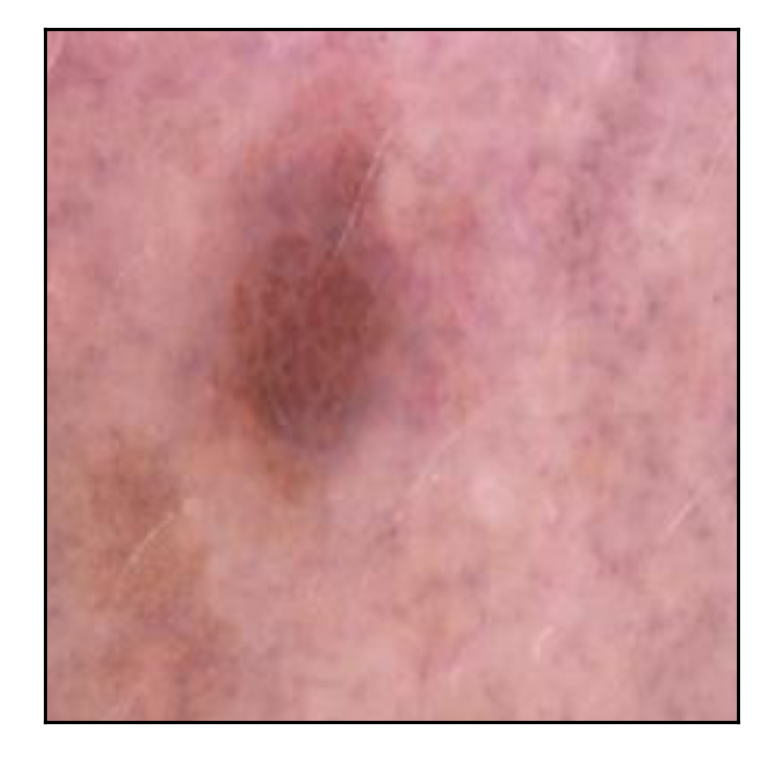}}
	\subfigure[]{\includegraphics[width=4.5cm]{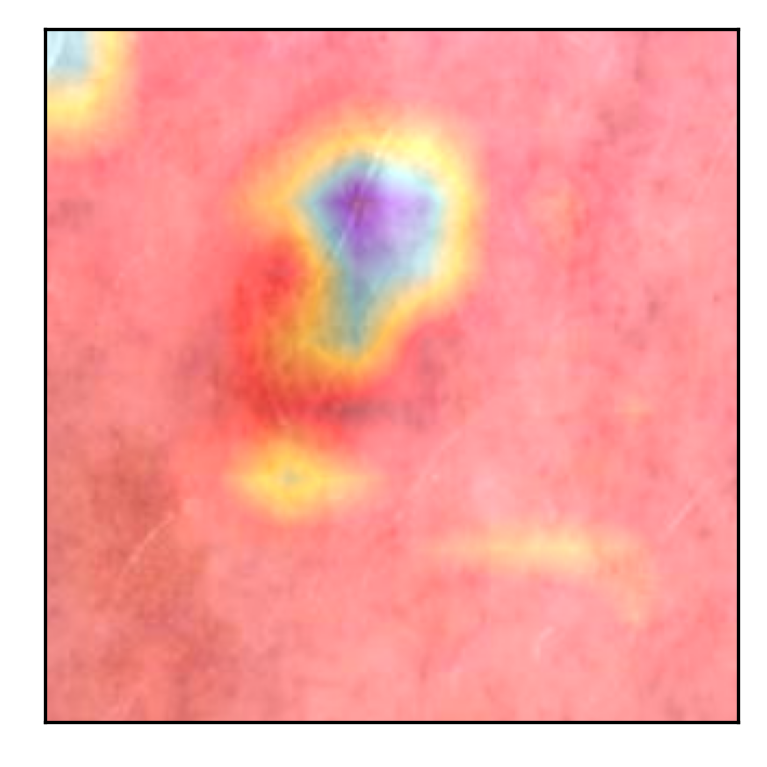}}
	\subfigure[]{\includegraphics[width=4.5cm]{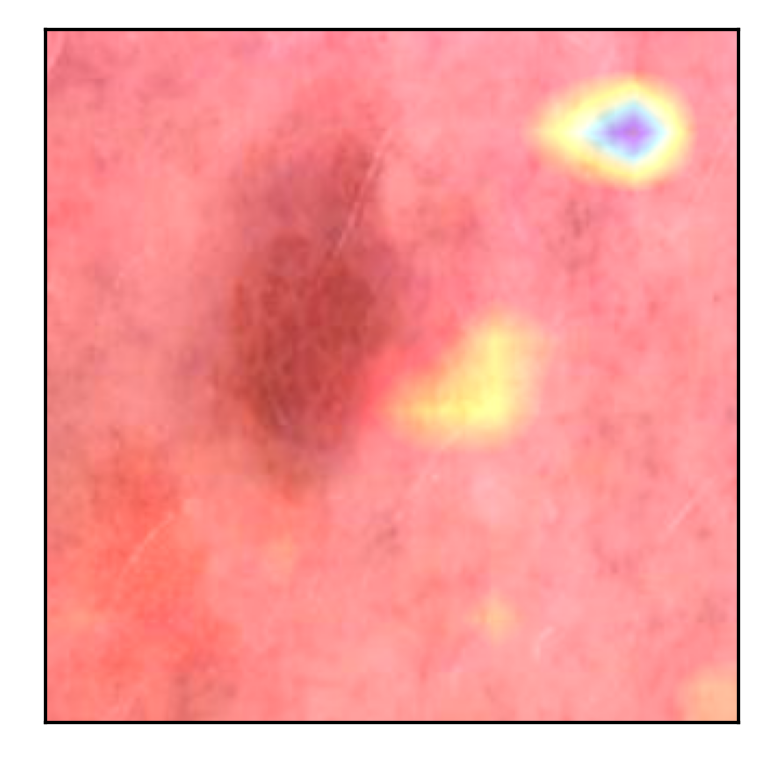}}\\
	\subfigure[]{\includegraphics[width=4.5cm]{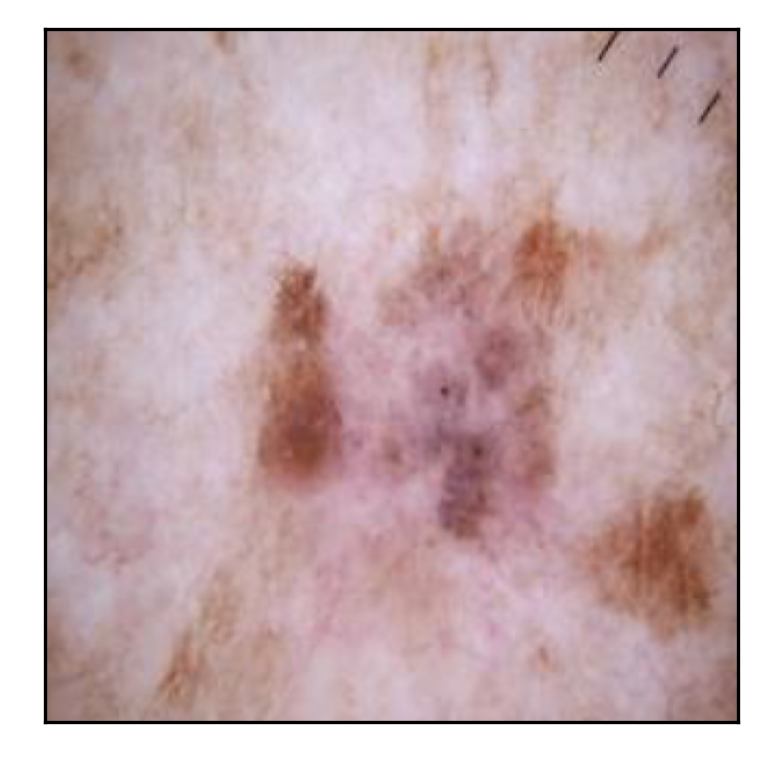}}
	\subfigure[]{\includegraphics[width=4.5cm]{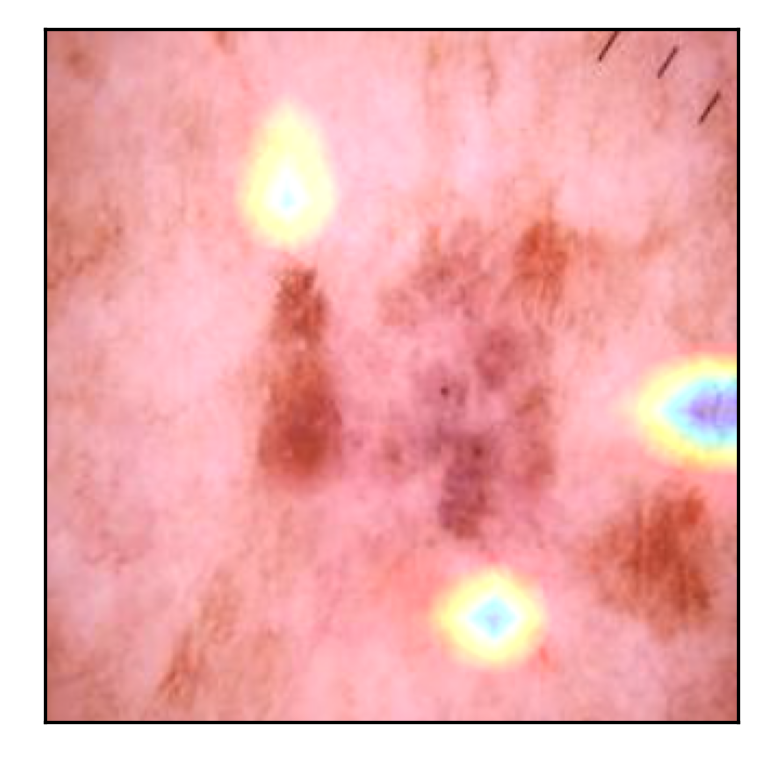}}
	\subfigure[]{\includegraphics[width=4.5cm]{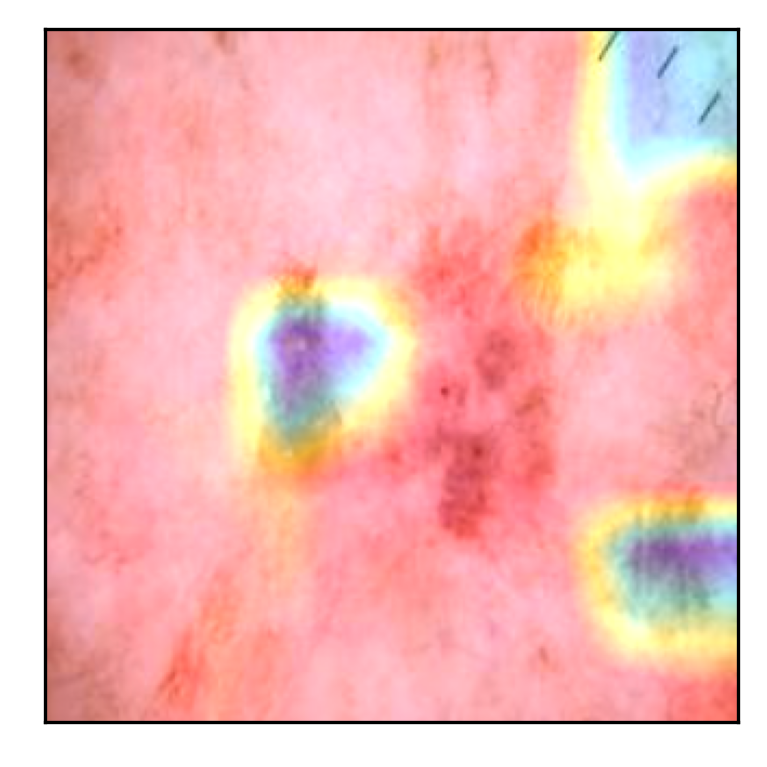}}		
	\caption{Application of the proposed framework on skin cancer dataset (a) Original input image$_1$ (b) Factual explanations on image$_1$ (c) Counterfactual explanations on image$_1$ (d) Original input image$_2$ (e) Factual explanations on image$_2$ (f) Counterfactual explanations on image$_2$}\label{Fig:Skin cancer use case}
\end{figure}

Figure \ref{Fig:Skin cancer use case} presents the results from the application of the proposed framework on two random images from skin cancer dataset, which belong to the same differences classes, i.e. the first image belongs to ``Benign'' class, while the second one belongs to ``Malignant'' class. The Siamese model's prediction was 0.781, which implies that the model predicts 
the similarity score: 21.9\% and that the input images belong to the different classes. Figures \ref{Fig:Skin cancer use case}(b) and \ref{Fig:Skin cancer use case}(e) present the factual explanations provided by Grad-CAM in order to identify the features, which impact the model's decisions. The interpretation of Figures \ref{Fig:Skin cancer use case}(b) suggests that the model focused on a small region on the skin while the interpretation of Figure \ref{Fig:Skin cancer use case}(e) reveals that the model was focused on the tumor area. This implies that model was focused on regions with dissimilar visual characteristics for predicting that the similarity score between the input images is considerably low. Furthermore, Figures \ref{Fig:Skin cancer use case}(c) and \ref{Fig:Skin cancer use case}(f) present the counterfactual explanations that demonstrate the region of each image in which the model would have been based for predicting that the images are similar. Clearly, the highlighted areas in both images possess no similar visual characteristics.

Notice that although the input images are looking similar for an non-expert human, tumor's characteristics such as texture, color and size are considered vital for separating benign from malignant cases. Therefore, a possible recommendation from this use case could be to use data augmentation based on transformation techniques (rotation, flip, crop, zoom, change the brightness, contrast and saturation, etc) in order to improve the model's performance.

\subsection{AirBnB dataset}

\begin{figure}[!ht]
	\centering
	\subfigure[]{\includegraphics[width=4.5cm]{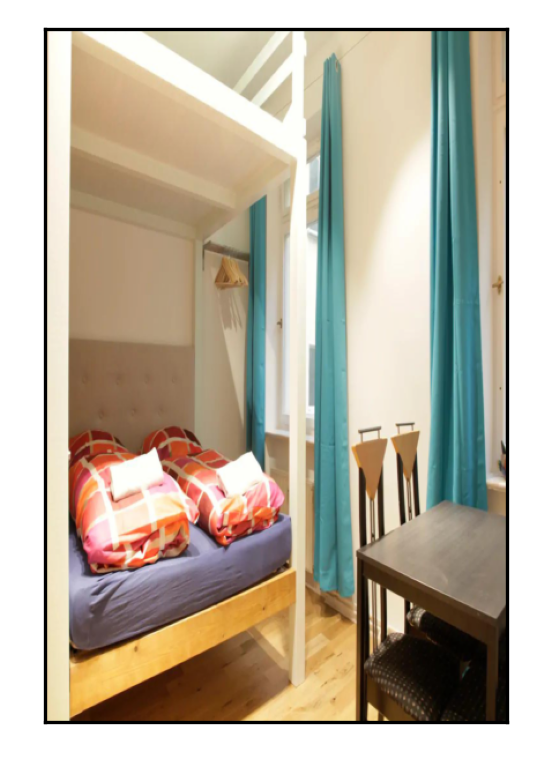}}
	\subfigure[]{\includegraphics[width=4.5cm]{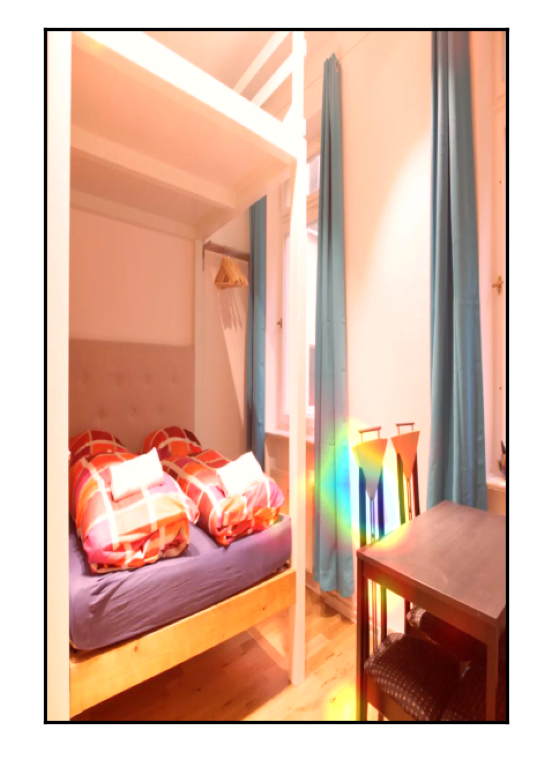}}
	\subfigure[]{\includegraphics[width=4.5cm]{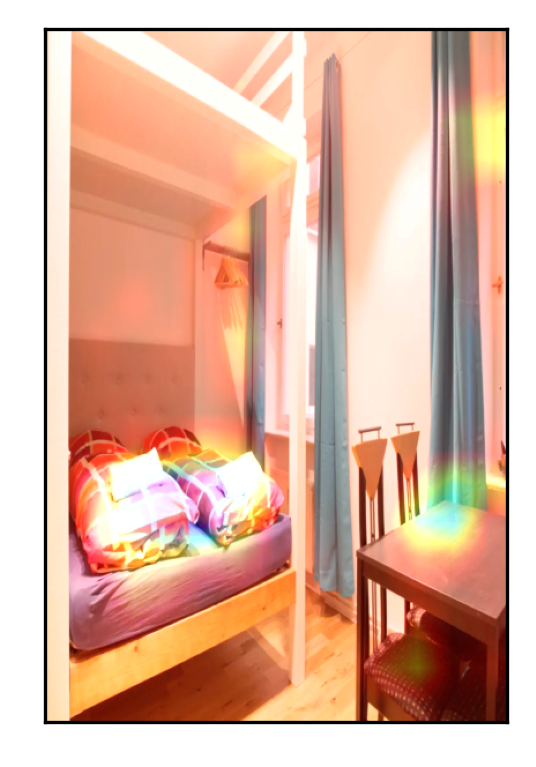}}\\
	\subfigure[]{\includegraphics[width=4.5cm]{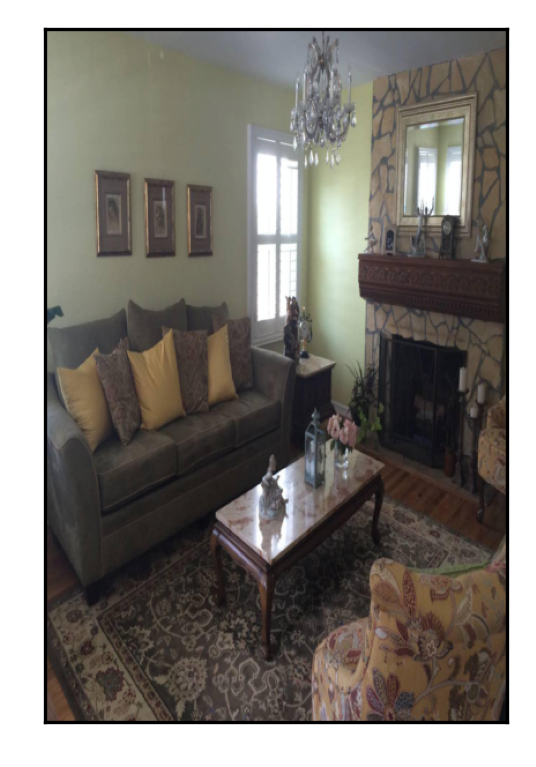}}
	\subfigure[]{\includegraphics[width=4.5cm]{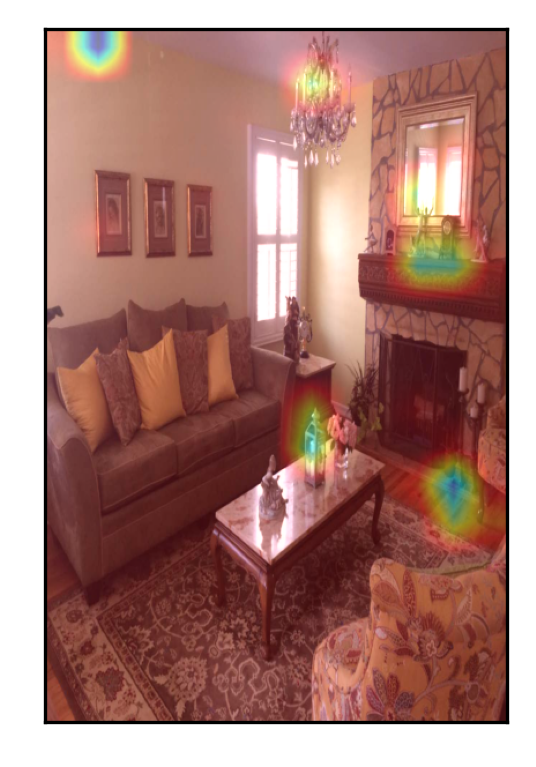}}
	\subfigure[]{\includegraphics[width=4.5cm]{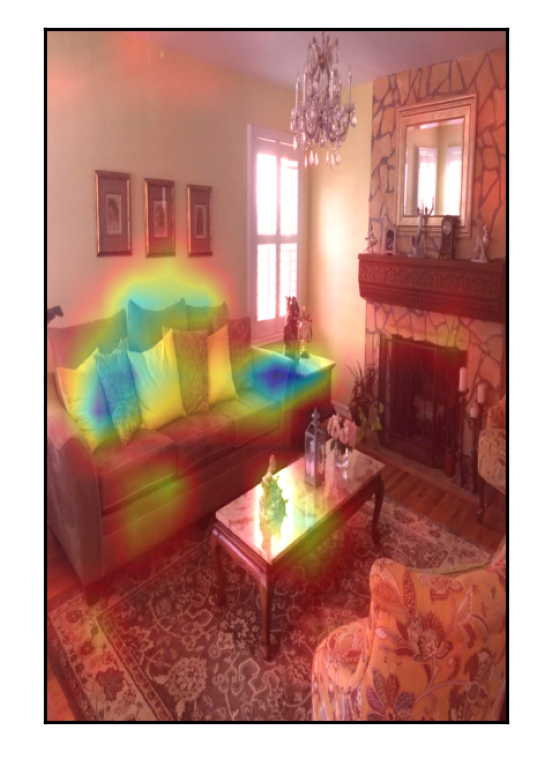}}		
	\caption{Application of the proposed framework on AirBnB dataset (a) Original input image$_1$ (b) Factual explanations on image$_1$ (c) Counterfactual explanations on image$_1$ (d) Original input image$_2$ (e) Factual explanations on image$_2$ (f) Counterfactual explanations on image$_2$}\label{Fig:AirBnB use case}
\end{figure}

Figure \ref{Fig:AirBnB use case} presents the results from the application of the proposed framework on two random images from AirBnB dataset, which belong to differences classes, i.e. the first image belongs to ``bedroom'' class, while the second one belongs to ``living-room'' class.
The Siamese model's prediction was 0.516, namely that the similarity score is 48.4\%, which suggests that the model predicts that the input images marginally belong to different classes. % LIVIERIS
Figures \ref{Fig:AirBnB use case}(b) and \ref{Fig:AirBnB use case}(e) present the factual explanations provided by Grad-CAM, which suggest that the model was focused on the chairs presented in the first image and on several items in the second image (such as lamps, fire-place and clock) to predict that the images are marginally dissimilar.

Since the model's prediction is not very confident; it is wise to study the counterfactual explanations to explore why the model was near to be confused. Figures \ref{Fig:AirBnB use case}(c) and \ref{Fig:AirBnB use case}(f) present the counterfactual explanations of both images, which suggest that the model, focused on the bed and sofa located in the first and second images, respectively as well as the tables presented in both images. This implies that the model was nearly confused since both images possess a common item (table) as well as two items which are visually similar (bed and sofa).

A possible recommendation for improving the model's performance could be to use advanced image processing techniques for item identification in order to assist the model of correlating the items and/or furniture, which belong to each room.

\subsection{Improving the Siamese model's performance}

In the rest of this section, we present an example of improving the performance of the Siamese model, through the conclusions and recommendations, which could be provided from the application of the proposed framework.
	
Firstly, we recall that in the use case scenario performed on Flowers dataset, we observed that by randomly selecting any pair of images, which belong to the same class, the Siamese model was focusing on the blossoms for making its decision (Figure \ref{Fig:Flower use case}). Hence, a possible recommendation for improving the model's performance could be that the model was based on identifying the blossoms in the input images for making its prediction; thus, a removal of characteristics such as stems, background, etc may improve the model's performance.
	
\begin{figure}[!ht]
		\centering
		\subfigure[]{\includegraphics[height=4cm]{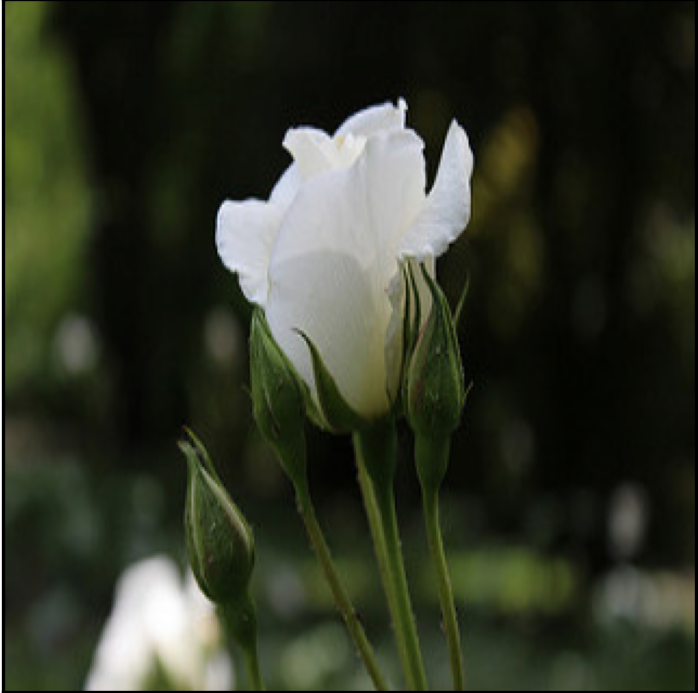}}\hspace{.4cm}
		\subfigure[]{\includegraphics[height=4cm]{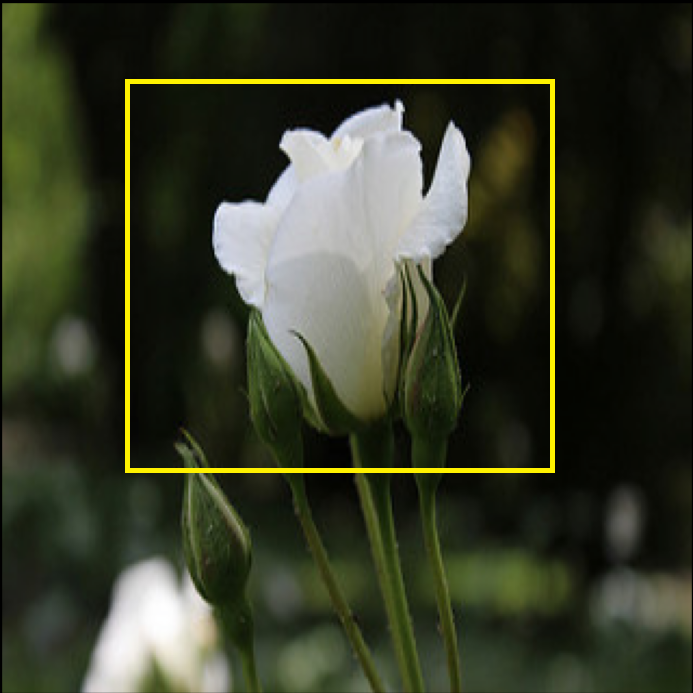}}\hspace{.4cm}
		\subfigure[]{\includegraphics[height=4cm]{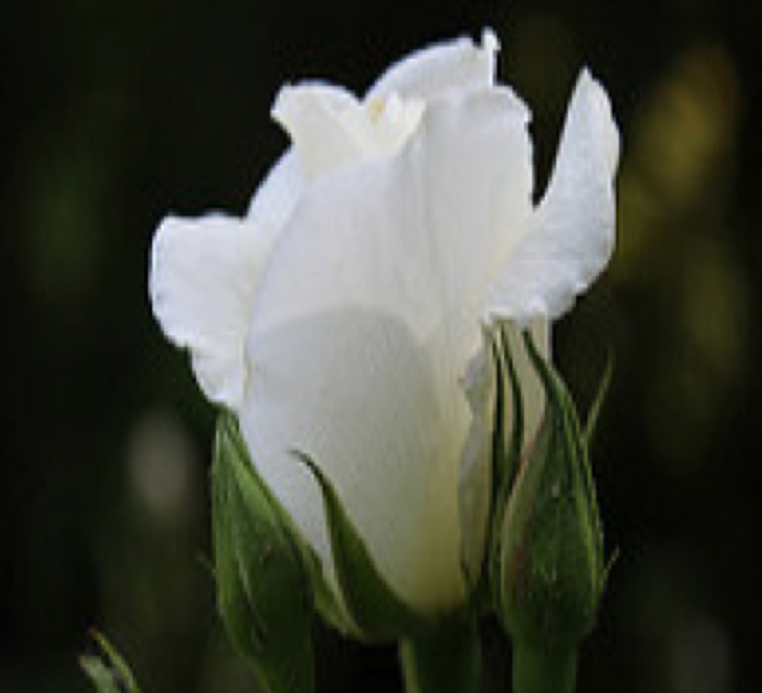}}
		\caption{(a) Original image (b) Bounding box (c) Cropped image}\label{Fig:Example}
	\end{figure}

To examine the effectiveness of this approach, we create a new dataset in which each figure is replaced with a bounding box containing the flower's blossom. For calculating the bounding boxes, for each image in the training data (anchor image) another image from the same class was randomly selected for calculating their similarity. In case, their predicted similarity by the model was > 80\%, then we calculate the anchor's image Grad-CAM heatmap. Based on the calculated heatmap, we utilized the methodology and implementation of Cui et al. \cite{cui2019chip}, for obtaining a bounding box, which contains the area which was mostly focused by the Siamese model for making its decision (i.e. flower's blossom). Along this line, in the newly created dataset, each image was replaced with the calculated bounding box. Figure \ref{Fig:Example} presents an example of the presented technique i.e. the original image, the bounding boxes of Grad-CAM and the cropped image.

Table \ref{Table:Evaluation} presents the performance of the Siamese model of identifying similar and dissimilar pairs of instances (a) trained with the original Flowers dataset (b) trained with the ``cropped'' Flowers dataset in which each image has been replaced with the identified bounding box using technique of Cui et al. \cite{cui2019chip}. The evaluation was performed on 432 pairs of similar and 432 pairs of dissimilar unseen images using accuracy, area under curve (AUC), precision and recall as performance metrics \cite{hossin2015review,livieris2019improving}. Clearly, we are able to conclude that the performance of the Siamese model was considerably increased relative to all performance metrics. In addition, the Siamese model achieved its top performance during the training process requiring less epochs in case it was trained with the ``cropped'' dataset.
	
	\begin{table}[!ht]
		\centering
		\begin{tabular}{l|ccccc}
			\toprule
			Dataset & Accuracy & AUC & Precision & Recall \\
			\midrule
			Original & 87.15\% &  0.872 & 0.890 & 0.872 \\
			``Cropped'' & 88.31\% &  0.883 & 0.900 & 0.880 \\
			\bottomrule
		\end{tabular}
		\caption{Siamese model's performance trained with the original and the ``cropped'' dataset}\label{Table:Evaluation}
	\end{table}
	
\begin{figure}[!ht]
	\centering
%	\subfigure[]{\includegraphics[width=12cm]{images/Example/Original.png}}
	\subfigure[]{\includegraphics[width=12cm]{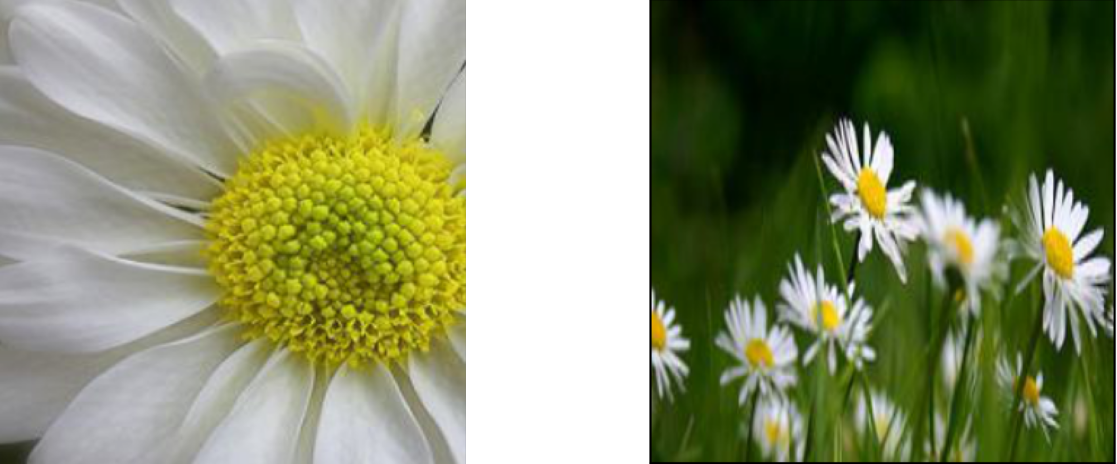}}	\caption{(a) Pair of images for the same class obtained from the original dataset (b) corresponding cropped images}\label{Fig:Example 2}
\end{figure}

Figure \ref{Fig:Example 2} presents two pairs of (similar) images from the same class (Daisy). The first pair contains images from the original Flowers dataset while the second pair contains the corresponding ``cropped'' images. 
	For the first pair the Siamese model predicted a similar score equal to 18\% while for the second pair the model predicted a similar score equal to 11\%.
	
Summarizing the previous discussion, we are able to conclude that the recommendation of removing characteristics such as stems, background and focusing on the blossoms 
considerable improved the quality of dataset.

\begin{figure}[!ht]
	\centering
	\subfigure[]{\includegraphics[width=12cm]{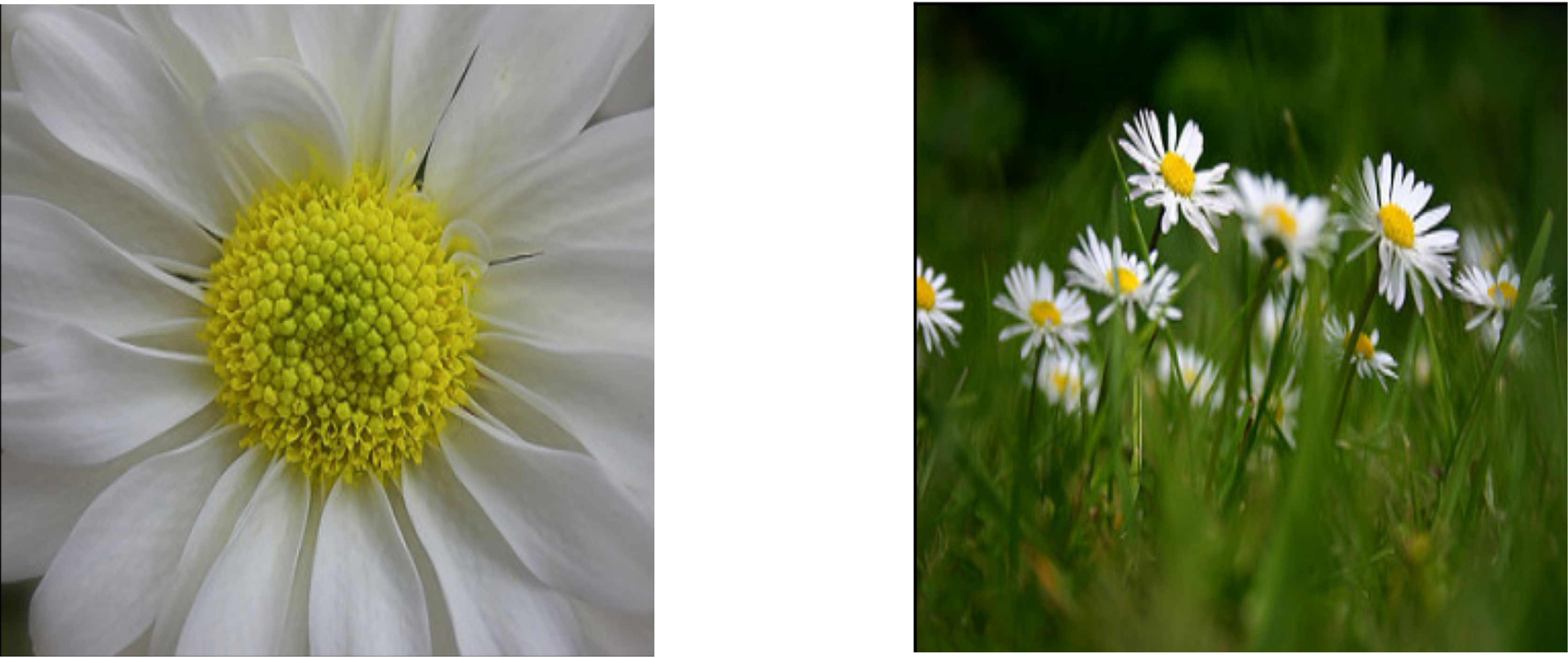}}
%	\subfigure[]{\includegraphics[width=12cm]{images/Example/Cropped.png}}	\caption{(a) Pair of images for the same class obtained from the original dataset (b) corresponding cropped images}\label{Fig:Example 2}
\end{figure}

\section{Discussion \& conclusions}\label{Sec:5}

The motivation for this research was to introduce the concept of explainable image similarity, for providing useful, interpretable and transparent insights into the underlying factors driving image relationships and comparisons.

In this modern deep learning area, models are becoming more and more complex, they can exhibit high accuracy but their lack of transparency and explainability becomes crucial for building trust and understanding. The context of explainable image similarity aims to bridge the gap between the black-box nature of sophisticated similarity models and human interpretability. The main goal is not only the development of models which are able to provide accurate similarity scores between pairs of images but also to offer insights into the specific features, patterns, or attributes that contribute to the computed similarity.  By offering interpretable explanations, explainable image similarity not only enhances the usability of similarity-based applications, such as image retrieval and recommendation systems, but also empowers users to comprehend the reasoning behind the models' decisions, ultimately fostering informed and confident decision-making.

For achieving this goal, we proposed a new framework by integrating Siamese networks together with Grad-CAM technique. The former are used for calculating the similarity between input images while the latter is used for visualizing and interpreting the decisions made by convolutional-based Siamese neural networks.
An attractive advantage of the proposed framework is that it is able to provide image similarity score along with visual intuitive explanations for its decisions. In addition, the proposed framework is able to evaluate bias on model's decisions as well as providing counterfactual explanations, highlighting the ability to "what if/model's decisions". 
The presented use cases scenarios included the application of the proposed framework on three similarity tasks from different application domains (two classification datasets and a few-shot learning dataset). Notice that the scope of this research was not to address a specific class of benchmarks i.e. few-shot learning benchmarks, one-shot learning benchmarks, etc but to provide human-meaningful explanations on similarity tasks. Clearly, the proposed framework can be easily applied on any image similarity tasks as well as few-shot/one-shot image classification tasks providing similarity scores along with visual explanations about its decisions. The use cases scenarios along with the provided comprehensive discussion highlighted the need for explainable image similarity and the useful conclusions and recommendations, which can be provided by its application.
Furthermore, we presented an example of improving the performance of the Siamese model for Flowers use case scenario, through the conclusions and recommendations provided from the application of the proposed framework. In more detail, the provided recommendations resulted in increasing the model's accuracy by 1.2\% and its prediction ability to identify similar images. For Skin cancer and AirBnB use case scenarios, the recommendations for improving the models performance were to use data augmentation based on transformation techniques (rotation, flip, crop, etc) and image processing techniques for item identification in order to correlate the items and/or furniture, which belong to each room, respectively.  Nevertheless, the former resulted in a minor improvement of the model's performance while the latter needs expert image processing and object identification techniques; hence, we decided to omit them.

It is worth mentioning that the proposed framework is based on the original Grad-CAM for providing visual explanations. Clearly, other state of the art techniques such as Grad-CAM++ \cite{chattopadhay2018grad}, XGrad-CAM \cite{fu2020axiom} and Score-Grad \cite{wang2020score}, can be easily adopted and incorporated. This can be considered as a limitation of this work; nevertheless, we should take into consideration that this was not the scope of this work. Another limitation can be considered the fact that the proposed framework uses a Siamese network with two input images. A possible extension could include the utilization of recent state-of-the-art models \cite{hu2022x} with more advanced and complex architectures as well as  the use of heatmap different saliency algorithms for heatmap calculation.
Some interesting works presented by RichardWebster et al. \cite{richardwebster2022doppelganger} and Hu et al. \cite{hu2023xaitk} 
used and proposed several algorithms for calculating saliency algorithms. An adoption of the proposed approach to their frameworks could provide useful conclusions from the factual and counterfactual explanations.

Our future work is concentrated on the application of the proposed framework on real-world image similarity benchmarks and its usage in conjunction with non post-hoc explainable techniques \cite{pintelas2020explainable,pintelas2021novel}.
Since the presented conclusions from the presented use case scenarios are quite encouraging, we intent to proceed with studying the accuracy performance impact on 
similarity tasks through the adoption of the proposed framework and the utilization of advanced image processing techniques.
Finally, another interesting idea could be the usage of advanced large language models for providing automated recommendations from the factual and/or counterfactual explanations \cite{peng2023instruction, topsakal2023creating}. 
Our expectation is that this research could be used as a reference for explainability frameworks, assisting decision-making by providing useful visual insights and offering customized assistance and recommendations on image similarity-related tasks.

%%%%%%%%%%%%%%%%%%%%%%%%%%%%%%%%%%
\bibliographystyle{unsrtnat}
\bibliography{bibliography}
%%%%%%%%%%%%%%%%%%%%%%%%%%%%%%%%%%

\end{document}